**A Deep Generative Model for Feasible and Diverse Population Synthesis**

**Eui-Jin Kim**
Research Fellow
Department of Civil and Environmental Engineering
National University of Singapore, Singapore, 117576
Email: euijin@nus.edu.sg

**Prateek Bansal**
Assistant Professor
Department of Civil and Environmental Engineering
National University of Singapore, Singapore, 117576
Email: prateekb@nus.edu.sg

Word Count: 6,998 words + 2 table (250 words per table) = 7,498 words

**ABSTRACT**

An ideal synthetic population, a key input to activity-based models, mimics the distribution of the individual- and household-level attributes in the actual population. Since the entire population's attributes are generally unavailable, household travel survey (HTS) samples are used for population synthesis. Synthesizing population by directly sampling from HTS ignores the attribute combinations that are unobserved in the HTS samples but exist in the population, called 'sampling zeros'. A deep generative model (DGM) can potentially synthesize the sampling zeros but at the expense of generating 'structural zeros' (i.e., the infeasible attribute combinations that do not exist in the population). This study proposes a novel method to minimize structural zeros while preserving sampling zeros. Two regularizations are devised to customize the training of the DGM and applied to a generative adversarial network (GAN) and a variational autoencoder (VAE). The adopted metrics for feasibility and diversity of the synthetic population indicate the capability of generating sampling and structural zeros – lower structural zeros and lower sampling zeros indicate the higher feasibility and the lower diversity, respectively. Results show that the proposed regularizations achieve considerable performance improvement in feasibility and diversity of the synthesized population over traditional models. The proposed VAE additionally generated 23.5% of the population ignored by the sample with 79.2% precision (i.e., 20.8% structural zeros rates), while the proposed GAN generated 18.3% of the ignored population with 89.0% precision. The proposed improvement in DGM generates a more feasible and diverse synthetic population, which is critical for the accuracy of an activity-based model.

**INTRODUCTION**

Activity-based models (ABMs) simulate and forecast daily activity tours of the population at an urban scale, which comprise multiple hierarchical dimensions of individual-level preferences in continuous time and space – when, where, for how long, in what sequence, and by which travel modes activities are performed (*1, 2*). The synthetic population is the key input to ABM, which aims to mimic the actual population's joint distribution of individual- and household-level attributes. Since attributes of the entire population are generally unavailable, regional household travel survey (HTS) samples (*1*), which cover about 1－5% of the population (*3*), are generally used in population synthesis. A traditional method for population synthesis is re-weighting, where the sampling weights are estimated from the aggregated population data (e.g., attribute distributions in each district), followed by simulating the population from the weighted samples (*1, 4, 5*). The main limitation of re-weighting is its inability to produce the attribute combinations that are not observed in the HTS samples but are present in the population.

Generative models (GMs) address this shortcoming of the re-weighting approach (*6, 7*). The GM first learns the joint probability distributions of the attributes followed by simulation, which enables the generation of synthetic agents with the attribute combinations that are not observed in the sample. The generated data from GM can be divided into four groups: (i) general sample, (ii) missing sample, (iii) sampling zero, and (iv) structural zero (*8, 9*). **Figure 1** illustrates these groups in a conceptual diagram. The general sample refers to the feasible generated data, while the missing sample is the feasible attribute combinations in the population and the HTS samples that could not be generated by the GM. The sampling zero refer to the attribute combinations included in the population but not included in the HTS sample. The capability of generating sampling zeros differentiates the GM from the re-weighting approach, but it comes at the expense of structural zeros (i.e., the infeasible attribute combinations, which do not exist in the population). The ideal synthetic population would include all sampling zeros and general samples without structural zeros and missing samples, but such data generation is not possible in practice due to an inherent trade-off. If the reliability of the generated data is of paramount importance, minimizing the structural zeros would be more critical than maximizing the sampling zeros.

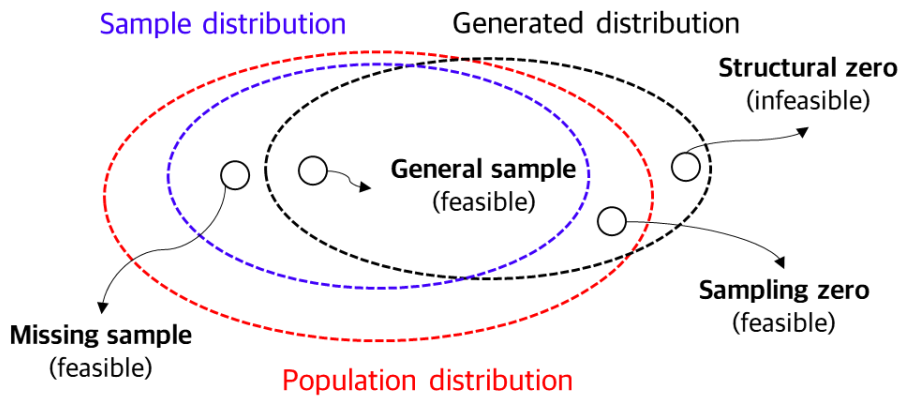

**Figure 1 Conceptual diagram of the general sample, missing sample, sampling zero, and structural zero**

Recent studies have applied a deep generative model (DGM) to population synthesis that incorporates a deep neural network into the GM (*8–11*). Even without customization, two widely-used DGMs, a variational autoencoder (VAE) and a generative adversarial network (GAN), achieved promising performance in reproducing the aggregated marginal and bivariate distribution of attributes. However, such metrics provide inadequate and misleading results in high-dimensional data (*12*), leading to overfitting the simplified training data distribution (*11*). Moreover, these metrics cannot evaluate the capability of generating sampling zeros which is the distinct strength of the GM. Recent studies in computer vision have evaluated the feasibility and diversity of DGMs to better understand the trade-off between structural and sampling zeros (*13, 14*). The lower structural zeros indicate the higher feasibility of the generated data, while the lower sampling zeros indicate the lower diversity. The feasibility and diversity metrics, however, have not yet been sufficiently discussed in the context of population synthesis.

This study aims to devise customized DGMs that minimize the structural zeros while preserving the sampling zeros and benchmark the performance of these novel DGMs based on feasibility and diversity metrics against traditional GAN and VAE, and advanced GMs such as Bayesian networks (BN). The central idea is to measure the distances from the generated data to the training sample distribution in the training procedure to evaluate whether the generated data is structural zero or sampling zero and use these measures to develop regularization terms to control the trade-off between the structural zero and sampling zero in the training of DGMs. In contrast to recent studies in the core DGM literature that have considered *generic regularizations* to enhance the diversity of the generated data (*15, 16*), this study focuses on the relationship between the structural and sampling zeros in devising regulation terms that is specific to the context of population synthesis. Such domain-knowledge-based regularization terms are expected to perform better than generic ones, as also illustrated by recent studies in the DGM-based choice models (*17, 18*).

In principle, the proposed model could be validated conditional on the availability of the difficult-to-obtain attributes of the entire population, but we circumvent this concern using a large-scale sample (> 1 million) of travel behavior data. Considering the sampling size of HTS data is below 100 thousand in previous studies (*8, 9, 11*), we assume the entire sample data as a hypothetical population (*h*-population) and a small portion of the data as a hypothetical sample (*h*-sample) for training. Thus, the *h*-population and *h*-sample are treated as population and sample in benchmarking the proposed DGMs against the existing approaches.

To the best of our knowledge, only (*8*) acknowledged the issue of sampling and structural zeros in population synthesis, but they did not achieve any model improvement to address this issue. Also, due to the limitation of the HTS samples containing around 75 thousand individuals, they defined the sampling zeros as the generated data that are not included in the training set (40%) but the test set (40%) and the structural zeros as the ones not observed in both training and test set. These definitions could lead to a bias in the structural zero and sampling zero rates because the distributions of the test set and training set are too similar to represent the relationship between sample data and full population data.

The contribution of this study is thus two-fold. First, beyond the vanilla GAN and VAE, we devise novel regularizations to enhance the feasibility and diversity of the synthetic population (i.e., minimize the structural zeros while preserving the sampling zeros). Second, we develop a rigorous model evaluation procedure using large-scale data and highlighting connections of novel metrics of diversity and feasibility with structural and sampling zero rates.

The next section describes the data and meaningful statistics of the *h*-sample and *h*-population. An explanation of GAN and VAE with the proposed regularizations is provided in the following section. Subsequently, the evaluation results are discussed. Conclusion and avenues of future work are summarized in the final section.

## DATA ACQUISITION
### Data preparation and assumptions
As discussed earlier, the issue of unavailability of the entire population is circumvented through the large-scale sample data with individual-level attributes of more than 1 million individuals, created by combining HTS data of South Korea conducted in 2010, 2016, and 2021. We assume the entire dataset as *h*-population and only 5% of them as *h*-sample (i.e., a sample of around 50 thousand individuals)(*3*). The *h*-sample can provide enough training data for DGM comparable to the previous studies that used HTS data for population synthesis (*8–11*). As expected and detailed in the next subsection, *h*-population and *h*-sample have substantial differences in diversity of attributes – the main requirement for a proper evaluation of the proposed method.

Table 1 provides descriptive statistics of *h*-population. It contains 13 individual attributes, of which all the numerical attributes are discretized into categorical attributes, and there are 70 categories in total. The categories of each attribute are adjusted to match the data from three different years. The data consisting of only categorical attributes make it possible to represent the data distribution with a finite number of attribute combinations. By doing so, we can identify whether the generated data is the general sample, missing sample, sampling zero, or structural zero.

**TABLE 1 Descriptive Statistics of the *h*-Population (N=1,066,319)**

| Attribute (Dimensions) | Category | Proportion (%) | Category | Proportion (%) |
|---|---|---|---|---|
| 1. Household income (6) | < 1 million won | 8.47 | 5 million – 10 million | 16.09 |
| | 1 million – 3 million | 39.46 | > 10 million won | 2.19 |
| | 3 million – 5 million | 33.78 | | |
| 2. Household car owner (2) | Yes | 83.91 | No | 16.19 |
| 3. Driver's license (2) | Yes | 60.13 | No | 39.87 |
| 4. Gender (2) | Male | 51.23 | Female | 48.77 |
| 5. Home type (6) | Apartment | 55.41 | Single house | 21.32 |
| | Villa | 12.09 | Dual purpose house | 0.82 |
| | Multi-family | 9.48 | Other | 0.89 |
| 6. Age (17) | 5 – 10 years | 4.96 | 51 – 55 years | 8.89 |
| | 11 – 15 years | 7.59 | 56 – 60 years | 7.38 |
| | 16 – 20 years | 7.48 | 61 – 65 years | 5.57 |
| | 21 – 25 years | 4.96 | 66 – 70 years | 4.27 |
| | 26 – 30 years | 6.08 | 71 – 75 years | 3.02 |
| | 31 – 35 years | 7.03 | 76 – 80 years | 2.23 |
| | 36 – 40 years | 9.42 | 81 – 85 years | 1.16 |
| | 41 – 45 years | 9.90 | 86 – 90 years | 0.42 |
| | 46 – 50 years | 9.67 | | |
| 7. Number of working days (4) | 5 days per week | 27.81 | 1 – 4 days per week | 10.05 |
| | 6 days per week | 17.33 | Inoccupation/non-regular | 44.82 |
| 8. Working types (9) | Student | 15.45 | Manager/Office | 11.54 |
| | Inoccupation/Housewife | 18.40 | Agriculture and fisher | 5.68 |
| | Experts | 11.07 | Simple labor | 12.31 |
| | Service | 15.69 | Others | 4.43 |
| | Sales | 5.44 | | |
| 9. Kid in the household (2) | Yes | 11.04 | No | 88.96 |
| 10. Number of households (7) | 1 | 7.56 | 5 | 9.67 |
| | 2 | 18.16 | 6 | 1.32 |
| | 3 | 25.27 | 7 | 0.14 |
| | 4 | 37.88 | | |
| 11. Travel mode of regular travel (6) | Car | 25.65 | Taxi | 0.31 |
| | Bike/Bicycle | 2.14 | Walking | 21.53 |
| | Public transportation | 22.49 | None | 27.87 |
| 12. Departure time of regular travel (4) | Peak | 56.38 | Others | 2.31 |
| | Non-Peak | 13.45 | None | 27.87 |
| 13. Students (4) | Kid | 0.58 | University | 4.75 |

| | Elementary/Middle/High | 18.01 | None | 76.67 |
|---|---|---|---|---|

*Note*: The 'regular travels' in the 11th and 12th attributes include working and non-working purposes such as commuting, going to school, and going to the senior citizen center.

**Sampling zeros according to a sampling rate**

Using the sample data weighted by aggregated population data as a simulation pool (*4, 5*) inevitably ignores the sampling zeros in the population synthesis. Therefore, we evaluate the degree of sampling zeros according to the sampling rate using the *h*-population and *h*-sample, as shown in **Figure 2**. First, we extract all the unique attribute combinations from the *h*-population and measure the number of unique combinations sampled from the *h*-sample, depicted as 'sampled combinations' (yellow line in the figure). The number of unique combinations in the *h*-population is 264,005, while that in the *h*-sample is only 30,837, indicating that the sampled unique combinations are only 11.7%. Second, we measure the ratio of data instances in the population whose attribute combinations are included in the sample. This 'sampled data instances' (red line in the figure) is a weighted 'sampled unique combinations' according to the number of observations in the *h*-population, which will be more generally defined as 'recall' in the later section. In the *h*-sample, 11.7% of sampled combinations represent 56.4% of *h*-population observations or data instances. This result implies that if the population synthesis is conducted by re-weighting procedure, the diversity of the synthetic population would be restricted to 56.4% of population data instances represented by 11.7% of attribute combinations.

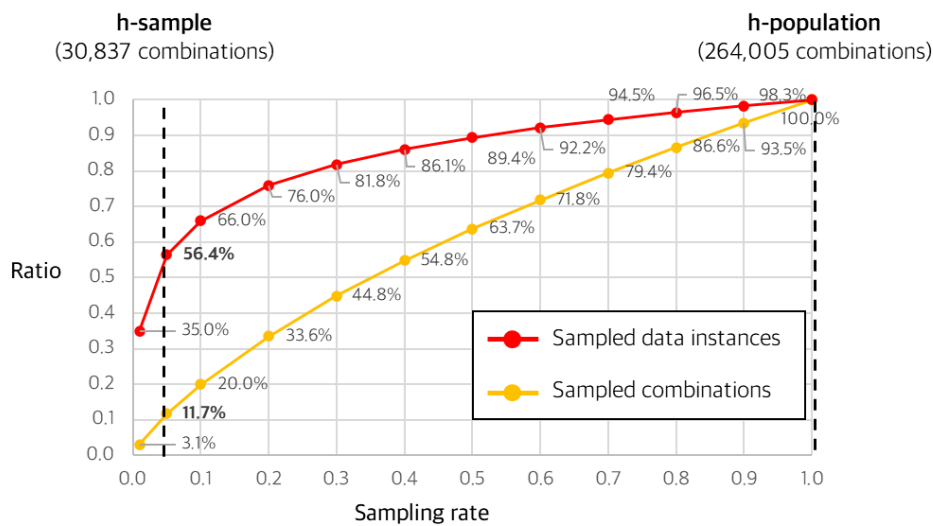

**Figure 2 The degree of sampling zeros according to the sampling rate**

**METHODOLOGIES**

**Hypothesis for sampling and structural zero**

We should characterize the sampling zero in training to distinguish it from the structural zero. This study hypothesizes that the sampling zeros may be located nearer the boundary of the sample distribution than the structural zeros because the sample boundary is a subset of the population. **Figure 3(a)** illustrates this hypothesis in the discrete space of two categorical attributes. In the discrete space, all the combinations of categorical attributes can be represented by finite combinations in which structural and sampling zeros can be identified. The three black circles in **Figure 3(a)** represent the general sample, sampling zero, and structural zero, and their distances to the nearest boundary of the sample distribution are 0, 1, and 3, which are aligned with the hypothesis. However, similar to the x-mark in the figure, some sampling zeros may be farther from the sample boundary than structural zeros due to two potential reasons. First, the shape of the probability density of the sample is different from those of the population. Second, the distance in the discrete space does not consider the contextual relationship between the attributes. For example, a person of $55 - 60$ years and $60 - 65$ years would be more different than other age groups due to retirement. Due to the second reason, this study considers another possibility to measure the distance. The embeddings are continuous vector representations of discrete data, and embedding neural networks are widely used to transform the high-dimensional discrete data into a lower-dimensional continuous vector (*19*). **Figure 3(b)** illustrates the general sample, sampling zero, and structural zero in the embedded continuous space. It is worth noting that the embedded space may also contain the structural zero against our hypothesis, as indicated by the x-mark in **Figure 3(b)** – a structural zero closer to sample distribution than sampling zero.

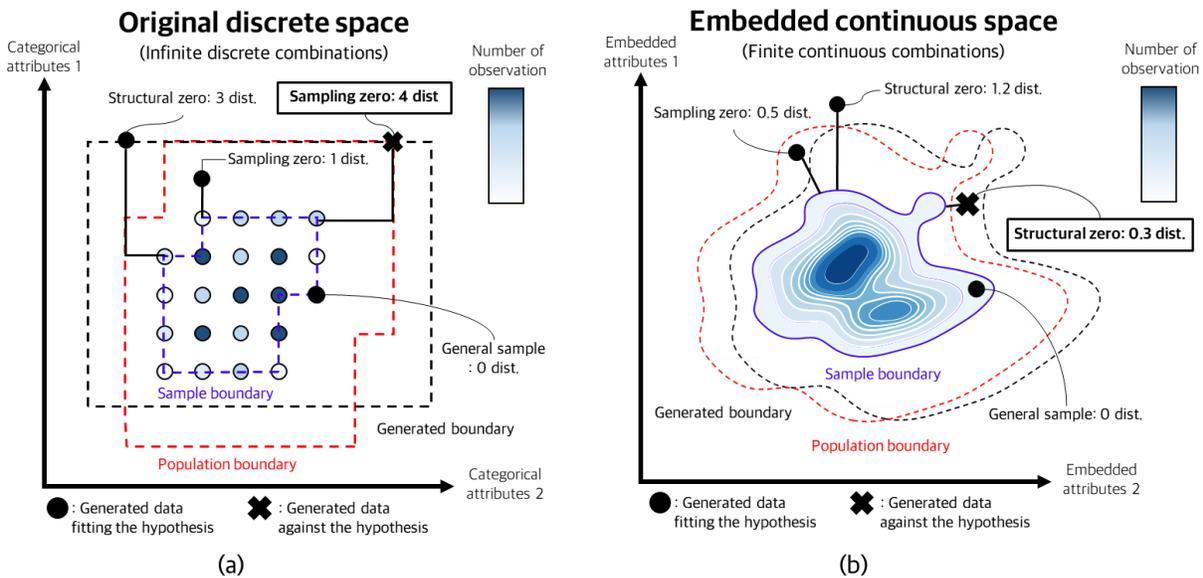

**Figure 3 Hypothesis for sampling and structural zero in the (a) original discrete space and (b) embedded continuous space.**

We employ the self-supervised embedding networks to transform the multi-categorical attributes into numeric attributes (*19*). The embedding network is trained to output a set of complete individual attributes given an incomplete set of attributes. In training, some of the input attributes are randomly chosen and masked, and these masked attributes are reused as target attributes to be estimated by neural networks. The self-supervised learning is repeatedly conducted until the performance for filling masked attributes is saturated. The contextual relationships among categorical attributes would be learned in filling the masked attributes. We use only the embedding parts of the neural network, which are depicted as a green box in **Figure 4**. Readers are referred to (*19*) for more details about the embedding networks.

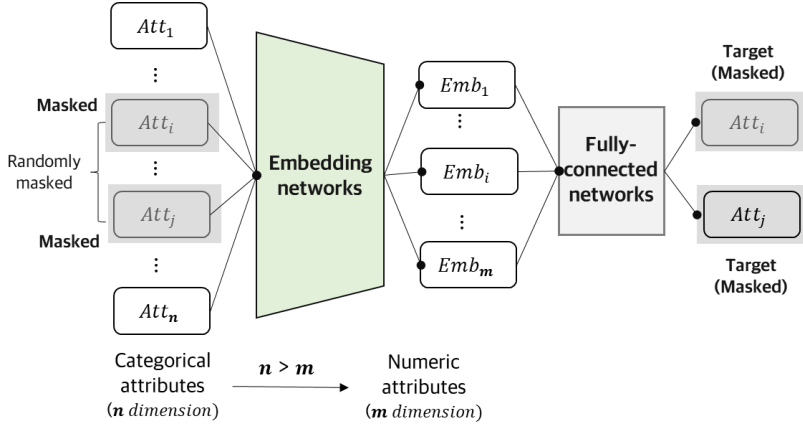

**Figure 4 Model structure of self-supervised embedding networks**

The distance between the two data vectors $(X_i, X_j)$ are measured by the Euclidean distance between two vectors as in **Equation 1**. In the discrete space, the sample distribution is mapped from the categorical vectors into the binary vectors because the generated data in training is represented as multiple categorical distributions (i.e., a sum of one for each categorical attribute). In the embedding space, two data points are represented as continuous vectors, so Euclidean distance is directly measured.

$$\text{Dist}(X_i, X_j) = \sqrt{(X_i - X_j)^2} \tag{1}$$

**Deep generative models (DGMs)**

The DGM approximates the joint probability distribution of the population's attributes, $P(\boldsymbol{X})$, using the sample's attributes, $\boldsymbol{X}^S$. Considering the superior performance of a GAN and a VAE for population synthesis (*8–11*), we adopt them as the two main models. The proposed regularizations to control the sampling and structural zeros are applied to GAN and VAE. Since the application of VAE and GAN is not a major contribution of this study, only high-level descriptions of them are provided in the following sections. More details about GAN and VAE are presented in (*20*) and (*21*), respectively.

*Generative adversarial network (GAN)*

The GAN estimates the joint probability distributions based on the simultaneous training of the generator and discriminator networks. These two networks play a min-max two-player game in which the generator tries to fool the discriminator by generating realistic data, and a discriminator tries to distinguish the generated data from the real data. The training continues until the equilibrium of a two-player game is achieved, where the discriminator no longer distinguishes between the generated and real data. Mathematically, the generator, $G(\boldsymbol{z}; \boldsymbol{\phi}_g)$, parameterized by $\boldsymbol{\phi}_g$ estimates the $P_{\boldsymbol{\phi}}(\tilde{\boldsymbol{X}})$, with a $\boldsymbol{z}$ sampled from a prior random distribution, $P(\boldsymbol{z})$ (e.g., multivariate normal distribution). The generator is a mapping function from $\boldsymbol{z}$ to $\boldsymbol{X}$ and is represented by neural networks. The parameterized discriminator, $D(\boldsymbol{X}; \boldsymbol{\phi}_d)$, outputs whether the attribute combination is the real data ($\boldsymbol{X}$) or the generated data ($\tilde{\boldsymbol{X}} = G(\boldsymbol{z})$) (e.g., the probability of $\boldsymbol{X}$ to be sampled from the real data). The parameters $\boldsymbol{\phi}_g$ and $\boldsymbol{\phi}_d$ are estimated using the value function denoted by **Equation 2.**

$$\min_{\boldsymbol{\phi}_g} \max_{\boldsymbol{\phi}_d} E_{X \sim P(X)}[log D(\boldsymbol{X})] + E_{z \sim P(z)}\left[\log\left(1 - D\left(G(\boldsymbol{z})\right)\right)\right]. \tag{2}$$

In each training epoch, the $D$ is trained to maximize the value function by imposing the higher $D(X)$ and lower $D(G(z))$ (i.e., the correct labels). Subsequently, the $G$ is trained to minimize $\log\left(1 - D(G(z))\right)$ by deceiving $D$ with realistic data.

The vanilla GAN (*21*) achieves equilibrium when the Jensen-Shannon divergence (JSD) between the generated distribution, $P_\phi(\hat{X})$, and the real distribution, $P(X)$, is minimized. However, when generating multi-categorical data, $P(X)$ is discrete on a $K$-dimensional set of simplex ($\{p_i \geq 0, \sum p_i = 1\}$) but $P_\phi(\hat{X})$ is continuous over this simplex set. This discrepancy leads to instability and saturation of equilibrium because it can cause the infinite JSD between $P_\phi(\hat{X})$ and $P(X)$ (*22*). Therefore, we employ a Wasserstein GAN (WGAN) with a gradient penalty (GP)(*22*). The WGAN is trained to minimize the Wasserstein distance (WD) between $P_\phi(\hat{X})$ and $P(X)$, instead of the JSD. The WD does not suffer from the issue of JSD since it is continuous and differentiable almost everywhere. The GP stabilizes the training by restricting the parameter search space of the discriminator. The loss function of vanilla WGAN-GP is sum of $\mathcal{L}_d$, $\mathcal{L}_g$, and $\mathcal{L}_{GP}$, which are defined in **Equations 3 to 6**.

$$\mathcal{L}_d = \frac{1}{m}\sum\nolimits_{i=1}^{m} -D(X_i) + D\big(G(z_i)\big). \tag{3}$$

$$\mathcal{L}_g = \frac{1}{m}\sum\nolimits_{i=1}^{m} -D\big(G(z_i)\big). \tag{4}$$

$$\mathcal{L}_{GP} = \frac{1}{m}\sum\nolimits_{i=1}^{m} \lambda\left(\left\|\nabla_{\tilde{X}_i}D\big(\tilde{X}_i\big)\right\|_2 - 1\right)^2. \tag{5}$$

$$\tilde{X}_i = \alpha\hat{X}_i + (1-\alpha)X_i, \ \alpha\sim\text{Uniform}[0,1]. \tag{6}$$

where $m$ is the size of the mini-batch sampled from the training data, $\|.\|_2$ is an Euclidian norm, $\hat{X}_i$ and $X_i$ are the generated and real data, respectively, $\tilde{X}_i$ is a weighted average of them, and $\lambda$ is the weight for the GP term. We used the generally recommended $\lambda$=10 (*22*). The $\mathcal{L}_d$ corresponds to maximizing the value function, and the $\mathcal{L}_{GP}$ regularizes it based on $D$'s gradients. The $\mathcal{L}_g$ indicates minimizing $\log\left(1 - D(G(z))\right)$ to train the $G$.

*Variational Autoencoder (VAE)*

A VAE consists of an encoder, $Q(X; \theta_e)$, and a decoder, $R(z; \theta_d)$. The encoder $Q$ maps the input data ($X$) into the parameters of latent prior distribution $P(z)$ (e.g., a vector of the mean and standard deviation of multivariate normal distribution), whereas the decoder $R$ generates the data ($\hat{X}$) mimicking $X$ from the $z$ sampled from $P(z)$. The decoder parameter $\theta_d$ is estimated together with the encoder parameter $\theta_e$ based on a reparameterization trick proposed by (*20*). Unlike the GAN's equilibrium-based training, the VAE is trained to minimize the discrepancy between the input and generated data. For stable training, the output of the encoder is regularized near the $P(z)$. We adopt a multivariate normal distribution as a $P(z)$. The loss function of VAE is sum of $\mathcal{L}_R$ and $\mathcal{L}_{KL}$, which are detailed in **Equations 7 and 8**:

$$\mathcal{L}_R = \frac{1}{m}\sum\nolimits_{i=1}^{m} X_i \log\hat{X}_i. \tag{7}$$

$$\mathcal{L}_{KL} = -\beta D_{KL}[Q(X)||P(z)]. \tag{8}$$

The first term indicates the cross-entropy between the generated and the input multi-categorical data. It induces minimizing the discrepancy between the input and generated data. The second term is the Kullback-Leibler (KL) divergence between the prior ($P(z)$) and estimated latent distribution that is the output of the encoder. The KL divergence measures the probabilistic distance between distributions, where $\beta$ was set to 1.0 in vanilla VAE (*20*). **Figure 5** depicts the training procedure of GAN and VAE with the loss functions.

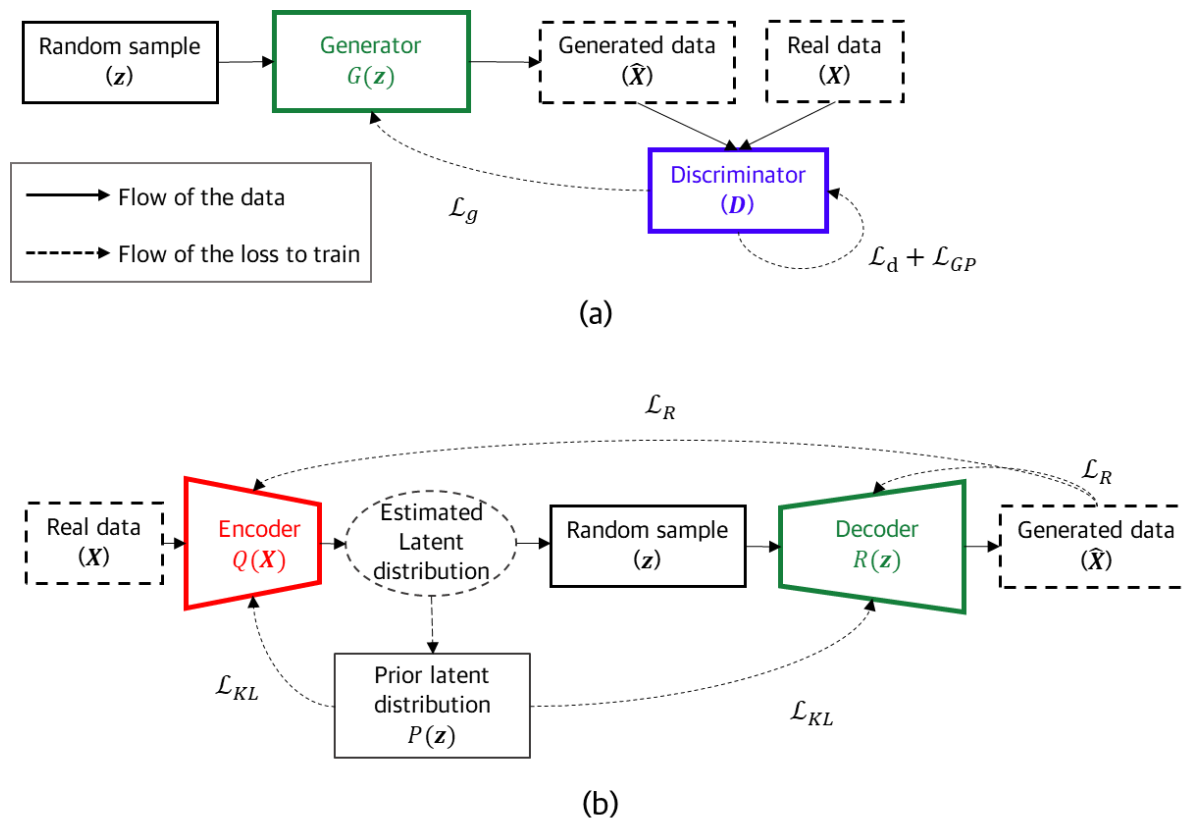

**Figure 5 The training procedure of (a) GAN and (b) VAE.**

**Regularizations considering the structural zeros and sampling zeros**

*Distance to the boundary of sample distribution*
This study hypothesizes that sampling zero's distance to the boundary of sample distribution is shorter than that of structural zero. To confirm this hypothesis, we synthesize the sampling and structural zeros from the vanilla VAE and WGAN, which are trained using *h*-samples. **Figure 6** shows the histogram of sampling and structural zeros' distance to the boundary. The results indicate that the distance to the boundary of sample distribution is a good indicator to distinguish the structural and sampling zeros in discrete and embedding spaces.

  The structural and sampling zeros generated by WGAN tend to be more clearly distinct than those generated by VAE. This difference comes from the inherent training procedure of WGAN and VAE (*13*). The VAE is trained by maximizing the likelihood, which leads to pursuing the intermediate characteristics of the data. On the other hand, the WGAN is trained by the equilibrium between the discriminator and generator, causing more distinct characteristics. Therefore, VAE tends to represent more smooth data distribution patterns than the WGAN, as also revealed in **Figure 6.**

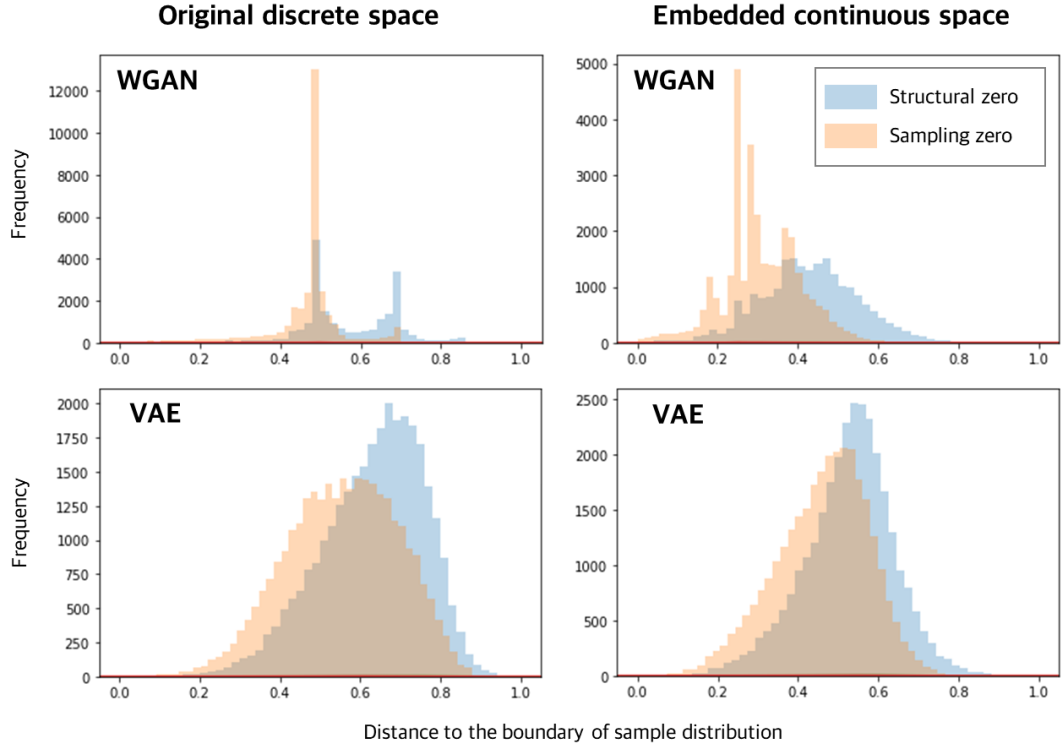

**Figure 6 Histograms of distance from sampling zeros and structural zeros to the boundary of sample distribution**

*Regularizations based on the distance to the sample distribution*

Based on the hypothesis verified in **Figure 6**, we devise the 'boundary distance regularization'($R_{BD}$) in **Equation 9**, which discourages the generation of sampling zero far from the sample boundary.

$$R_{BD}(\widehat{X}, X^S) = \frac{1}{M}\sum_{j=1}^{M} \min_{i\in\{1:N\}, j\in\{1:M\}} (Dist(\widehat{X}_j, X_i^S)) \tag{9}$$

where the $\widehat{X}_j$ and $X_i^S$ are generated and sample data. The $N$ and $M$ are the training data and mini-batch sizes, respectively. The $R_{BD}$ calculates the nearest distance of each generated data to $N$ training data and averages them for $M$ generated data. **Figure 7** helps clarify $R_{BD}$ on the simplified data space. The black x-marks in the figure indicate the structural zero filtered out by $R_{BD}$. The gap between the structural zeros filtered by $R_{BD}$ and the sample distribution is calibrated by the training data. The $R_{BD}$ is expected to reduce the structural zero but can also falsely remove the sampling zero (i.e., red circle in the figure).

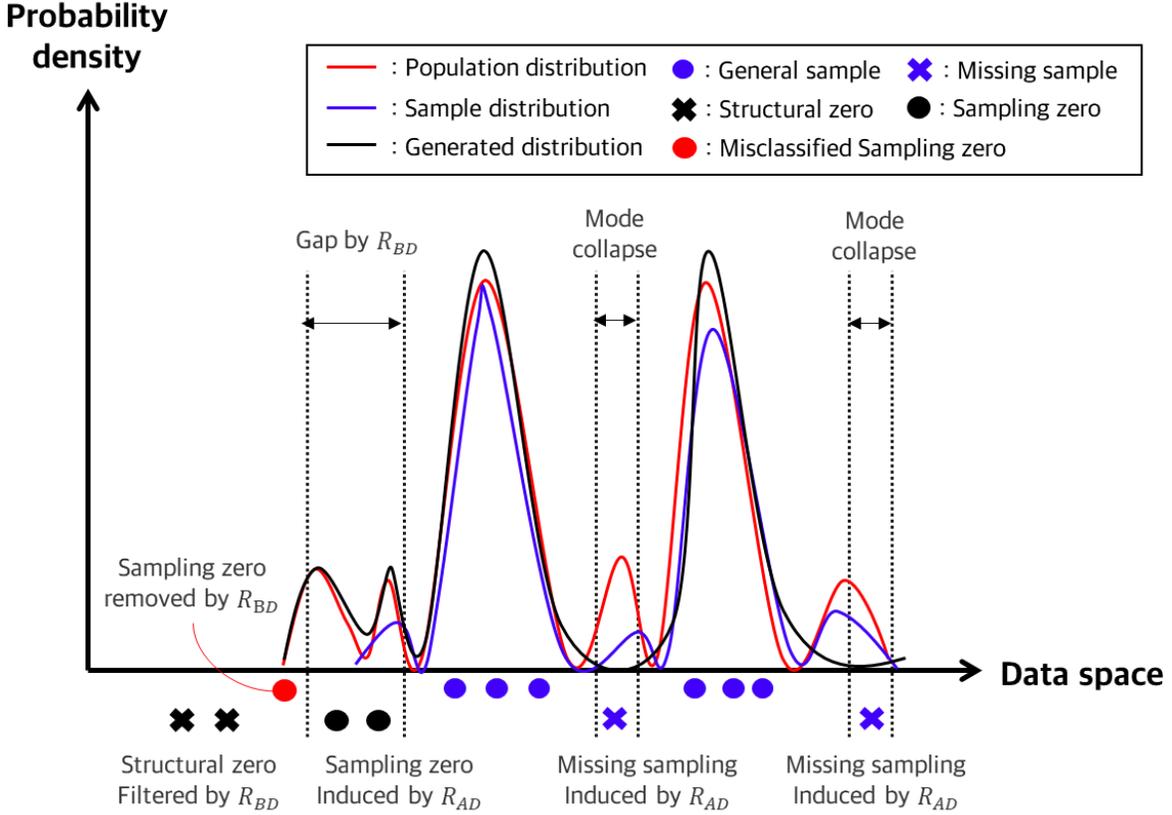

**Figure 7 The generated data considered by $R_{BD}$ and $R_{AD}$ on the simplified data space**

Although the DGM tries to reproduce the sample distribution, there can be missing samples that exist in the sample distribution but do not exist in the generated distribution (See **Figure 1**). These missing samples are caused by the 'mode collapse,' a tendency of the DGM to ignore minor attribute combinations. The mode collapse is more prevalent in GAN than in VAE due to its equilibrium-based training (*16*). Since the missing samples are included in the sample distribution, the distance to the boundary of the missing sample becomes zero. To induce the generation of the missing sample, we devise 'average distance regularization' ($R_{AD}$) denoted by **Equation 10.**

$$R_{AD} = -\frac{1}{NM} \sum_{j=1}^{M} \sum_{i=1}^{N} Dist(\hat{X}_j, X_i^S) \tag{10}$$

The $R_{AD}$ computes the average for the 'average distance to the sample distribution' of the $M$ generated data. As shown in **Figure 7**, the missing sample and sampling zero tend to have a long average distance to the sample; thus, the $R_{AD}$, that is negative, encourages the generation of missing samples as well as sampling zero.

*Variants of the DGM*

This study proposes two regularizations, $R_{BD}$ and $R_{AD}$. The $R_{BD}$ prevents generating the structural zero, whereas the $R_{AD}$ induces generating the sampling zero and missing sample. We apply both regularizations to VAE and GAN, and evaluate their impact. The final loss functions of GAN and VAE adopting both regularizations are shown in **Equations 11 and 12,** respectively. The regularization weights are calibrated with the metrics described in the following section.

$$\mathcal{L}_{GAN} = \mathcal{L}_{d} + \mathcal{L}_{G} + \mathcal{L}_{GP} + \gamma_{BD}^{GAN} R_{BD} + \gamma_{AD}^{GAN} R_{AD} \tag{11}$$

$$\mathcal{L}_{VAE} = \mathcal{L}_{R} + \mathcal{L}_{KL} + \gamma_{BD}^{VAE} R_{BD} + \gamma_{AD}^{VAE} R_{AD} \tag{12}$$

**Evaluation metrics**

*Distributional similarity*

Intuitively, the DGM can be evaluated by the distributional similarity between the population distribution ($X$) and the generated distribution ($\hat{X}$). However, comparing the high-dimensional distributions is computationally expensive and fails to provide consistent results (*12*). Alternatively, some researchers evaluate the simplified distributional similarity, such as the aggregated marginal and bivariate distributions (*6, 9, 10, 23*). Although these metrics risk overfitting to marginal and bivariate distribution (*11*), they can be a schematic indicator of whether the generated synthetic data by the DGM is reasonable or not. We adopt standardized root mean square error (SRMSE) to evaluate the distributional similarity of marginal and bivariate distribution (*9*). **Equation 13** denotes the SRMSE for bivariate distributions.

$$\text{SRMSE}(\boldsymbol{\pi}, \hat{\boldsymbol{\pi}}) = \frac{\text{RMSE}(\boldsymbol{\pi}, \hat{\boldsymbol{\pi}})}{\overline{\boldsymbol{\pi}}} = \frac{\sqrt{\sum_{(k,k')} \left(\pi_{(k,k')} - \hat{\pi}_{(k,k')}\right)^2 / N_b}}{\sum_{(k,k')} \pi_{(k,k')} / N_b}, \tag{13}$$

where $\boldsymbol{\pi}$ and $\hat{\boldsymbol{\pi}}$ are categorical distributions of $h$-population and the generated data, respectively. $N_b$ is the total number of category combinations. For the bivariate distribution of categorical variables $k \neq k'$, $\boldsymbol{\pi} = \{\boldsymbol{\pi}(X_1, X_2), \dots, \boldsymbol{\pi}(X_k, X_{k'})\}$ becomes the vector of $\binom{K}{2}$ bivariate combinations of categorical variables. In the case of the marginal distribution, the vector $\boldsymbol{\pi} = \{\pi(X_1), \dots, \pi(X_K)\}$ becomes the vector of the marginal distribution of $K$ categorical variables.

*Feasibility and diversity*

This study evaluates the proposed DGM based on feasibility and diversity of the generated data because the sampling zeros and structural zeros are closely related to them. Such measures have been proposed to replace the distributional similarity by recent studies in computer vision (*13, 14, 24*). Diversity refers to the degree to which the generated data captures the population variations. Whereas feasibility indicates how well the generated data resembles the population data. The feasibility and diversity are evaluated by precision and recall, respectively. This study proposes modified definitions of the precision and recall for population synthesis based on the $h$-population ($X$) and generated data ($\hat{X}$), as shown in **Equations 14 and 15.** Since precision and recall have a trade-off relationship, we calculate the F1 score, indicating the overall quality as shown in **Equation 16:**

$$\text{Precision} = \frac{1}{M} \sum_{j=1}^{M} 1_{\hat{X}_j \in X} \tag{14}$$

$$\text{Recall} = \frac{1}{N} \sum_{i=1}^{N} 1_{X_i \in \hat{X}} \tag{15}$$

$$\text{F1 score} = \frac{2 \times \text{Precision} \times \text{Recall}}{\text{Nrecision} + \text{Recall}} \tag{16}$$

The $1(\cdot)$ is an indicator function for counting. **Figure 8** depicts how feasibility and diversity are connected to precision and recall. The precision measures the ratio of generated data included in the feasible $h$-population, that is one minus structural zero rates. The recall measures the ratio of $h$-population whose combinations are included in the generated data, which indicate the population's diversity captured by the generated data. The recall is also proportional to the sampling zero.

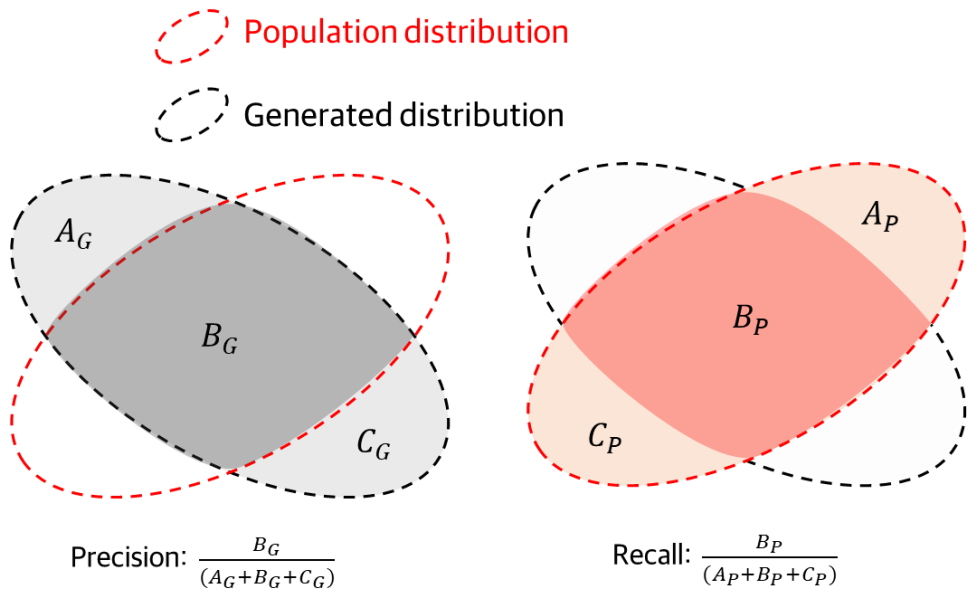

**Figure 8 Conceptual diagram of precision and recall for evaluating generative model**

## EMPIRICAL APPLICATION

### Model evaluation results

**Table 2** shows the evaluation results for the proposed DGM compared to vanilla ones, BN, and re-weighting. The re-weighting fits the marginal distributions of attributes in the sample to those in the population. Since the *h*-sample has marginal distributions similar to the *h*-population, we generate re-weighted samples directly from the *h*-sample (i.e., equal weights). The BN is a GM that decomposes the joint distribution into a set of partial conditional distributions to learn the data generating process efficiently. The structure of partial conditional distributions is determined by heuristic methods considering its complexity and training performance. We apply a hill-climbing algorithm (*25*) to determine the structure of BN. Readers are referred to (*7*) for more details about BN for population synthesis. Regarding DGM, we separately evaluate the effects of two regularization terms for both GAN and VAE. The regularization terms are calculated in the discrete and embedding dimensions. We evaluate the model performance based on distributional similarity, diversity, feasibility, and overall quality. Since diversity and feasibility have a trade-off relationship, we calibrate the hyperparameters based on overall quality, i.e., the model variants in **Table 2** are selected based on the F1 score. Also, the diversity of the DGMs varies according to the number of generated samples. Considering the application to population synthesis, each DGM generates the data with the *h*-population's size.

We extensively tune hyperparameters of the DGMs since it affects the quality of the training. A grid search is conducted for a comparable degree of tuning for each model variant. Specifically, we tune the learning rate, activation function, batch size, the number of layers and neurons, latent distribution dimension, and weights for the proposed regularizations. **Figure 9** summarizes the calibrated architecture and hyperparameters of the WGAN and VAE.

**TABLE 2 Evaluation Results of Generated Data with Size of *h*-population.**

| Method | | | Distributional Similarity | | Diversity | | Feasibility | Overall Quality |
|---|---|---|---|---|---|---|---|---|
| Model | Space | Regularization | Marg. SRMSE | Bivar. SRMSE | # of comb* | Recall | Precision | F1 score |
| Re-weighting | | | 0.008 | 0.020 | 30,387 | 56.4% | 100.0% | 72.1% |
| BN | | | **0.009** | 0.084 | 303,723 | 78.4% | 73.7% | 76.0% |
| VAE | - | Vanilla | 0.055 | 0.127 | 355,277 | 81.0% | 71.8% | 76.1% |
| | Discrete (Dis) | $R_{BD}$ | 0.095 | 0.218 | 265,875 | 79.9% | 79.2% | 79.5% |
| | | $R_{AD}$ | 0.057 | 0.132 | 312,506 | **82.1%** | 76.1% | 79.0% |
| | | $R_{BD}$ & $R_{AD}$ | 0.079 | 0.173 | 329,097 | 82.0% | 73.6% | 77.6% |
| | Embedded (Emb) | $R_{BD}$ | 0.088 | 0.208 | 289,377 | 80.2% | 76.6% | 78.4% |
| | | $R_{AD}$ | 0.060 | 0.140 | 334,569 | **82.3%** | 74.1% | 78.0% |
| | | $R_{BD}$ & $R_{AD}$ | 0.050 | 0.116 | 318,731 | 82.0% | 75.4% | 78.6% |
| WGAN | - | Vanilla | 0.022 | 0.064 | 279,336 | 80.2% | 79.7% | 79.9% |
| | Discrete (Dis) | $R_{BD}$ | 0.036 | 0.094 | 155,586 | 74.7% | **89.0%** | **81.2%** |
| | | $R_{AD}$ | **0.016** | **0.048** | 273,622 | 81.2% | 80.4% | 80.8% |
| | | $R_{BD}$ & $R_{AD}$ | 0.043 | 0.106 | 152,031 | 74.1% | **89.2%** | 81.0% |
| | Embedded (Emb) | $R_{BD}$ | 0.023 | 0.076 | 225,408 | 77.7% | 84.6% | **81.0%** |
| | | $R_{AD}$ | 0.020 | **0.059** | 276,012 | 81.3% | 80.3% | 80.8% |
| | | $R_{BD}$ & $R_{AD}$ | 0.024 | 0.072 | 236,238 | 78.1% | 83.0% | 80.8% |

*Note*: **Bold** font indicates the best and second-best models for each metric except the re-weighting. # of comb. indicates the number of unique combinations of the generated data.
*: The number of combinations of population data is 264,005.

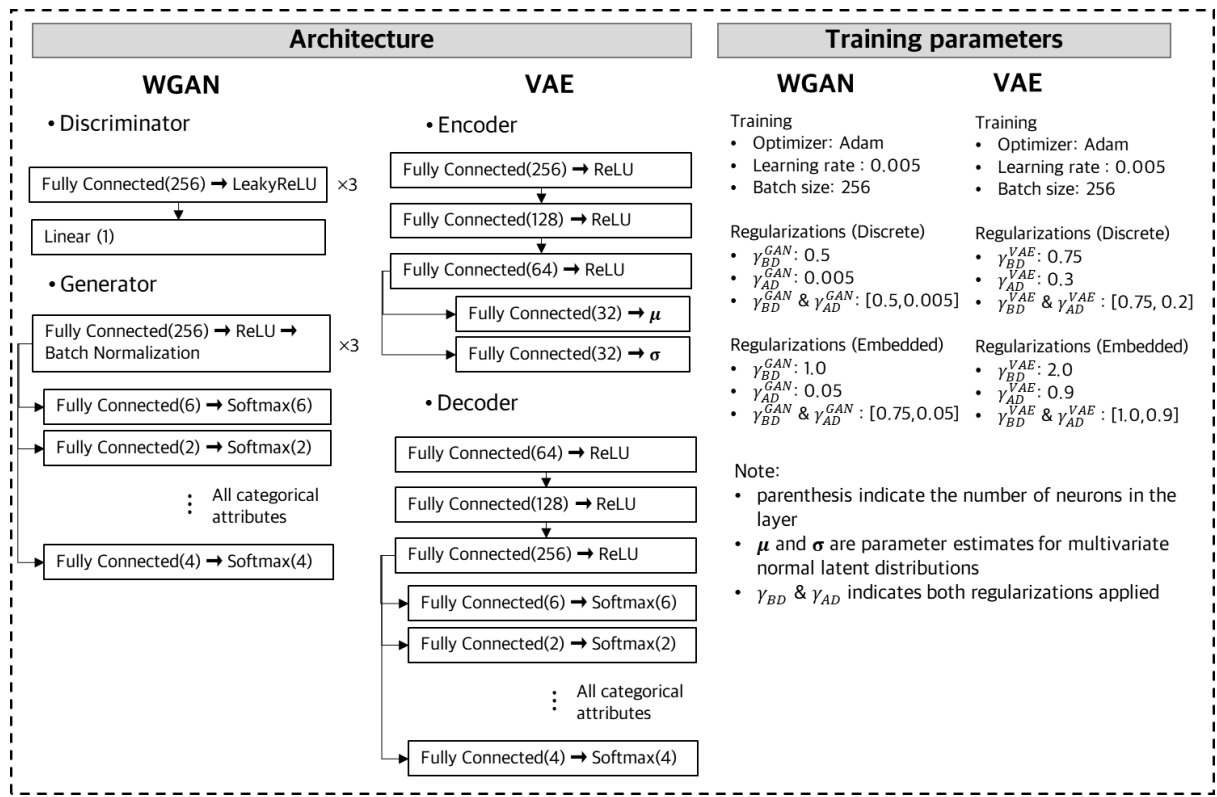

**Figure 9 Calibrated hyperparameters of WGAN and VAE**

*Distributional similarity*

The distributional similarity is measured by the SRMSE between marginal and bivariate distributions of generated and population data. Although it is vulnerable to overfitting the aggregated marginal and bivariate distributions (*11*), the distributional similarity is an intuitive indicator for checking whether the DGMs have been properly trained. Four main implications are derived from the results of distributional similarity**.** First, the BN shows the best performance for marginal SRMSE among GMs (even close to re-weighting), while the 'WGAN-Dis-$R_{AD}$' is the best for bivariate distribution. Considering the well-established superior performance of the BN in terms of distributional similarity (*7, 9, 11*), the superiority of WGAN for bivariate distribution is a promising result. Second, all the WGAN variants outperform their VAE counterparts. The WGAN tends to generate more similar and realistic data than the VAE at the expense of ignoring some minor attribute combinations (*13*). Such a tendency would be advantageous in estimating the aggregated distributions. Third, the $R_{AD}$ regularization in discrete and embedded spaces increases the distributional similarity of vanilla WGAN but decreases those of VAE. This result suggests that the regularizations work differently in WGAN and VAE. Lastly, the $R_{BD}$ regularization increases the SRMSE for all cases, indicating that the SRMSE cannot reflect aspects related to the structural zero. This result implies that future studies should not entirely rely on distributional similarity measures while evaluating the performance of methods to synthesize the population. **Figure 10** depicts the distributional similarity of 'WGAN-Emb-$R_{BD}$ & $R_{AD}$', demonstrating a good training of DGMs.

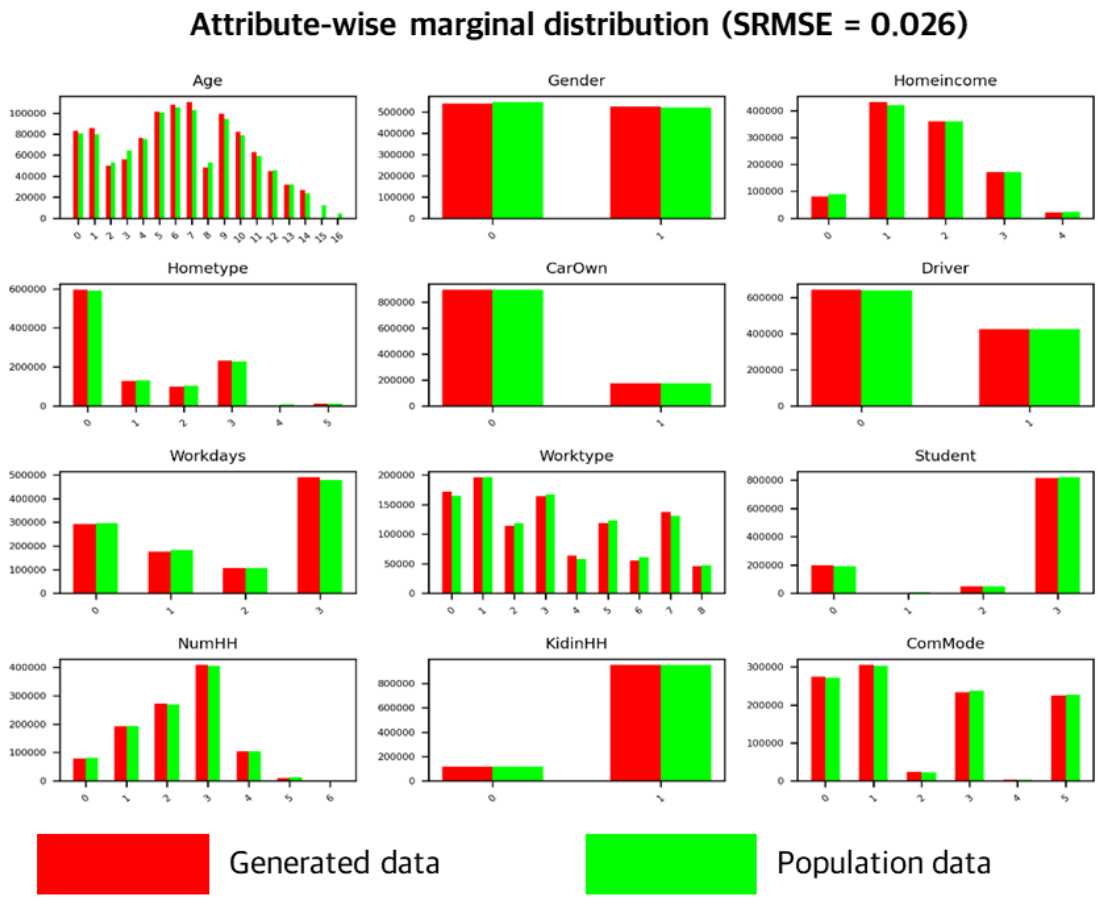

**Figure 10 Distributional similarity between the *h*-population and generated samples (WGAN-Emb-$R_{BD}$ & $R_{AD}$)**

*Diversity measured by recall*

The diversity measured by recall indicates the proportion of population data whose attribute combinations are covered by the generated data. The diversity is closely related to the sampling zero, a distinct superiority of GMs over re-weighting. We obtain five insightful patterns while analyzing the results of diversity measures. First, as expected, the re-weighting exhibits poor diversity, but the proposed DGM significantly enhances the diversity. Compared to the re-weighting's recall of 56.4%, the VAE-Emb-$R_{AD}$ achieves the recall of 82.3%, i.e., it additionally considers the 25.9 of the population ignored by the *h*-sample in the population synthesis. Second, the VAE variants marginally outperform the WGAN variants in terms of diversity. Such a pattern was also observed in previous studies in other domains (*13*). The VAE tends to smoothen the probability density distributions, leading to more diverse but fuzzy data generation. In contrast, the GAN tends to generate realistic data while ignoring some minor modes. The property of VAE pursuing the intermediate features is rather helpful in generating diverse data. Third, the $R_{AD}$ regularization consistently improves the diversity of GAN and VAE in both embedding and discrete spaces, but the improvements are marginal. An increase in diversity conflicts with the goals of GAN and VAE to mimic the training sample data. Therefore, $R_{AD}$ regularization's influence would inevitably be limited not to undermine the DGM's original objectives. Forth, the improvements from embedding and discrete spaces are almost identical, indicating that capturing the attribute relationships by embedding distance does not improve diversity. Lastly, the number of generated attribute combinations is partially related to the recall but is not proportional. The VAE variants generate more unique attribute combinations than WGAN variants, which is aligned with their higher recall. Meanwhile, the VAE-Emb-$R_{AD}$ exhibits higher recall than vanilla VAE but had fewer attribute combinations. This tendency is common regardless of the DGMs and data spaces. This result implies that the $R_{AD}$ regularization improves the diversity by recovering minor modes rather than rare attribute combinations. It supports our hypothesis the $R_{AD}$ regularization encourages generating the missing samples as well as sampling zero.

By definition, the recall increases in proportion to the number of generated data points. Since the available sampling rate is different for each researcher, the number of target-generated data would be different. For example, if the sampling rate is 10%, we only need to generate data 10 times the sample size. **Figure 11** represents the increasing recall patterns of WGAN-Emb-$R_{AD}$ and VAE-Emb-$R_{AD}$, compared to the vanilla ones. Initially, WGAN outperforms the VAE at 2 times the sample size (i.e., 50% sampling); however, both VAE-Emb-$R_{AD}$ and VAE achieve better performance at 20 times the sample size (5% sampling) than their counterparts of WGAN. In other words, the superior diversity of VAE over WGAN becomes more prominent for small sampling rates.

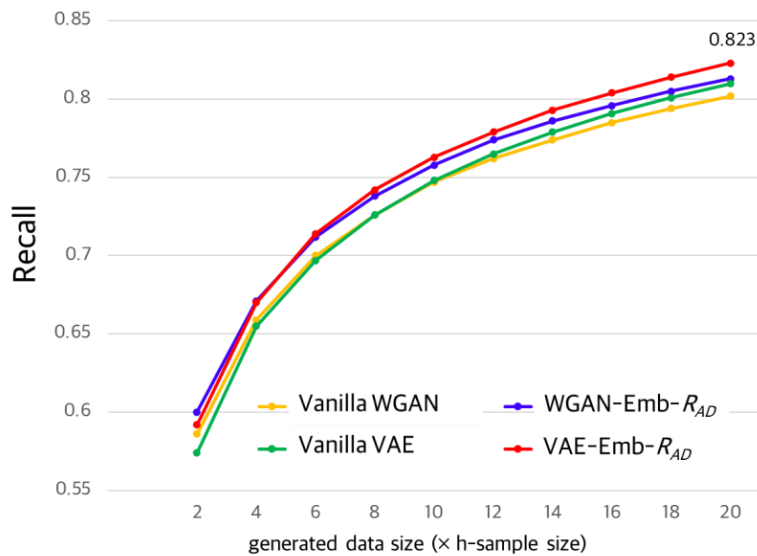

**Figure 11 The changes in diversity of DGMs according to the number of generated data**

*Feasibility measured by precision*

The feasibility measured by precision indicates the proportion of generated data whose attributes exist in the population data. Minimizing the infeasible synthetic population is important to secure the reliability of population synthesis but at the cost of sacrificing the sampling zero. We obtain three meaningful trends in the results of the feasibility measure. First, the WGAN with strength in producing realistic data shows higher precision than other models. Second, the proposed $R_{BD}$ regularization enhances the precision of WGAN and VAE by 9.3% and 7.4% compared to their vanilla counterparts. These results verify that the $R_{BD}$ regularization helps identify the sampling and structural zeros in the training procedure. Third, the precision enhancement from $R_{BD}$ regularization is more prominent in the discrete space than in the embedding space, but the accompanying loss of recall is less in the embedding space. The embedding space refines the boundary distance because both sample and generated data are defined in the continuous dimension. Therefore, a moderate point with less precision gain and less recall loss is likely to be selected as the optimal point.

*Overall quality*

Since feasibility and diversity have a trade-off relationship, we consider the overall quality (indicated by F1 score) as the main metric for the model evaluation. The above evaluations show that VAE has strengths in diversity and WGAN in feasibility. Whereas the best-performed WGAN achieves an 81.2% F1 score, which covers the attribute combinations of the 74.7% of population data with 89.0% precision. The best-performed VAE has a F1 score of 79.5% with 79.9% recall and 79.2% precision. These results indicate that although the WGAN exhibits better overall quality than VAE, the selection of the method depends on the goal of the decision-maker. Even though the re-weighting has 100% precision, its poor diversity (56.4%) makes it perform lower overall quality than the WGAN and VAE.

Since the $R_{BD}$ and $R_{AD}$ regularizations have opposite goals, they are expected to complement each other in training. Despite the extensive hyperparameter calibrations, combination of $R_{BD}$ & $R_{AD}$ regularities did not attain the best performance in any DGM specification. Therefore, based on the trade-off between feasibility and diversity, it would be better to find the optimal weight of each regularization than to find the weight combination of two regularizations.

*Sensitivity analysis of regularization terms*

For a deeper understanding of the importance of regularizations in training, **Figure 12** provides the sensitivity analysis according to the weights, $\gamma_{BD}$ and $\gamma_{AD}$. The WGAN with $R_{BD}$ and $R_{AD}$ regularizations in the discrete space are analyzed. The higher the $\gamma_{BD}$, the higher the precision and it monotonically increases while sacrificing the recall. The $R_{BD}$ regularization works as a boundary condition to filter out the structural zeros, and this boundary becomes narrower as the weights increase; thus, structural zeros decrease more and more. This result indicates that the proposed $R_{BD}$ regularization can control the trade-off between feasibility and diversity to accomplish an optimal balanced performance.

Meanwhile, the $R_{AD}$ regularization achieves a peak F1 score at an intermediate weight, which corresponds to the peak precision and slightly lower recall than the maximum. The $R_{AD}$ regularization is designed to encourage the generation of not only sampling zero but also missing samples, leading to improvement in both recall and precision. However, if $R_{AD}$ regularization is overweighted (e.g., 0.007 in this case), the generated data deviate from the population distribution beyond improving diversity, leading to reductions in precision and recall. Therefore, the optimal weight of $R_{AD}$ regularization should be carefully derived to ensure optimal performance in terms of the F1 score.

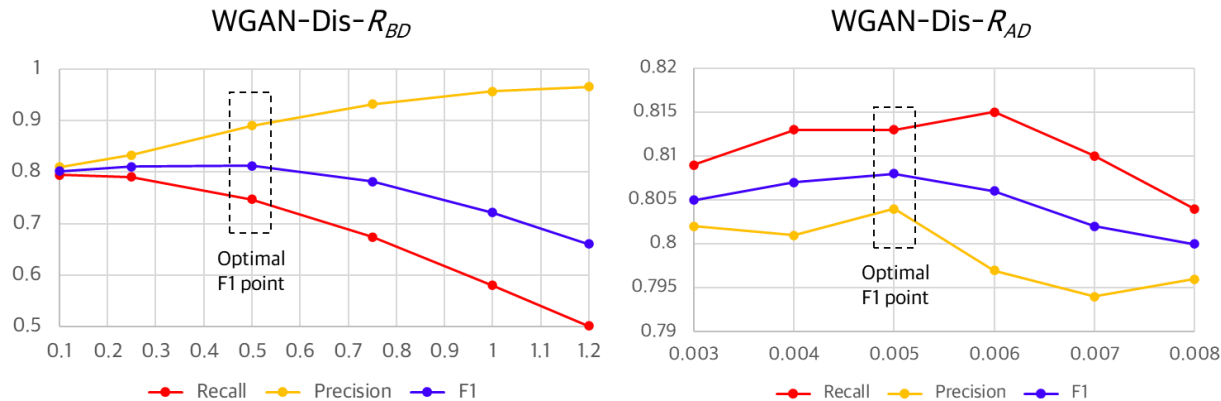

**Figure 12 Sensitivity analysis of the weights of regularizations on WGAN performance.**

## CONCLUSIONS

Population synthesis inevitably depends on small-scale household travel survey datasets due to the absence of entire population data, leading to several missing attribute combinations in the synthetic population that exist in the actual population. Recent advancements in deep generative models (DGMs) – variational auto encoder (VAE) and generative adversarial networks (GANs) — have helped in generating such missing attribute combinations (i.e., sampling zeros), but at the expense of generating infeasible attribute combinations (i.e., structural zeros). This study proposes novel regularization terms in the training of DGMs to reduce the structural zeros while preserving the sampling zeros. Novel evaluation metrics regarding feasibility (i.e., the proportion of the generated data whose attributes exist in the population data) and diversity (i.e., the proportion of the population whose attributes are covered by the generated data) are devised to test the extent of sampling and structural zeros. We benchmark the modified DGMs against vanilla DGMs and other traditional methods such as reweighting and Bayesian network (BN) based on these metrics.

This study offers several novel insights for applications of DGMs in population synthesis. First, distributional similarity measures could be misleading. The diversity and feasibility of the generated data should be computed. Second, the DGM provides overall better data quality than re-weighting by enhancing the diversity while sacrificing feasibility. Third, VAE has strengths in improving diversity and WGAN in improving feasibility of the generated data. We suggest selecting them depending on the decision maker's goal. Fourth, the proposed regularization terms enhances the DGM by controlling the trade-off between feasibility and diversity. The $R_{BD}$ regularization improves the feasibility by removing the structural zeros, and the calibrated weight controls the loss of diversity. The $R_{AD}$ regularization improves both feasibility and diversity by encouraging the generation of the missing samples and sampling zeros when the proper weight is imposed. The effect of regularization is more pronounced on precision compared to recall of the generated data – the $R_{BD}$ regularization enhances the precision of WGAN and VAE by 9.3% and 7.4% compared to their vanilla counterparts, but improvement in recall in both DGMs due to the $R_{AD}$ regularization is below 1.5%, perhaps because an increase in diversity conflicts with the goals of GAN and VAE to mimic the training sample data.

Emerging technologies such as big data, computation power, and artificial intelligence will make the activity-based models (ABMs) more disaggregate (*26*), which requires more diverse but feasible data at an individual level. The proposed method can meet these needs and control the trade-off between diversity and feasibility depending on the modeling goals. Also, the rights of minorities and fairness have started gaining attention in travel demand modeling (*27*), but machine learning inherently focuses on model-fit while ignoring the minority groups (*28*). The proposed regularizations and evaluation metrics contribute to addressing these issues.

With a vision to integrate population synthesis and activity plan generation using the DGM, the proposed improvements and evaluation metrics open up an important avenue for future research.

Specifically, the applications of DGMs in activity scheduling (*10*) are likely to suffer from similar issues related to the lack of diversity and feasibility. Generating activity schedules even requires much more modeling complexity than population synthesis due to the high-dimension spatiotemporal correlation between different dimensions of activity chains. Leveraging large-scale trip-chain data and effective regularizations could be a promising direction to generate diverse yet feasible activity patterns on an urban scale.

## ACKNOWLEDGMENTS

Prateek Bansal acknowledges financial support from the Presidential Young Professorship Grant.

## AUTHOR CONTRIBUTIONS

The authors confirm contribution to the paper as follows: study conception and design: E.-J. Kim, P. Bansal; data collection: E.-J. Kim; analysis and interpretation of results: E.-J. Kim, P. Bansal; draft manuscript preparation: E.-J. Kim, P. Bansal. All authors reviewed the results and approved the final version of the manuscript.

## REFERENCES

1.  Castiglione, J., M. Bradley, and J. Gliebe. *Activity-Based Travel Demand Models: A Primer*. Transportation Research Board, Washington, D.C., 2014.

2.  Axhausen, K. W., A. Horni, and K. Nagel. *The Multi-Agent Transport Simulation MATSim*. Ubiquity Press, 2016.

3.  Habib, K. N., W. El-Assi, and T. Lin. How Large Is Too Large? A Review of the Issues Related to Sample Size Requirements of Regional Household Travel Surveys with a Case Study on the Greater Toronto and Hamilton Area (GTHA). 2020.

4.  Rich, J. Large-Scale Spatial Population Synthesis for Denmark. *European Transport Research Review*, Vol. 10, No. 2, 2018. https://doi.org/10.1186/s12544-018-0336-2.

5.  Hörl, S., and M. Balac. Synthetic Population and Travel Demand for Paris and Île-de-France Based on Open and Publicly Available Data. *Transportation Research Part C: Emerging Technologies*, Vol. 130, No. July, 2021, p. 103291. https://doi.org/10.1016/j.trc.2021.103291.

6.  Farooq, B., M. Bierlaire, R. Hurtubia, and G. Flötteröd. Simulation Based Population Synthesis. *Transportation Research Part B: Methodological*, Vol. 58, 2013, pp. 243–263. https://doi.org/10.1016/j.trb.2013.09.012.

7.  Sun, L., and A. Erath. A Bayesian Network Approach for Population Synthesis. *Transportation Research Part C: Emerging Technologies*, Vol. 61, 2015, pp. 49–62. https://doi.org/10.1016/j.trc.2015.10.010.

8.  Garrido, S., S. S. Borysov, F. C. Pereira, and J. Rich. Prediction of Rare Feature Combinations in Population Synthesis: Application of Deep Generative Modelling. *Transportation Research Part C: Emerging Technologies*, Vol. 120, No. September, 2020, p. 102787. https://doi.org/10.1016/j.trc.2020.102787.

9.  Borysov, S. S., J. Rich, and F. C. Pereira. How to Generate Micro-Agents? A Deep Generative Modeling Approach to Population Synthesis. *Transportation Research Part C: Emerging Technologies*, Vol. 106, No. July, 2019, pp. 73–97. https://doi.org/10.1016/j.trc.2019.07.006.

10. Badu-Marfo, G., B. Farooq, and Z. Patterson. Composite Travel Generative Adversarial Networks for Tabular and Sequential Population Synthesis. *IEEE Transactions on Intelligent Transportation Systems*, 2022. https://doi.org/10.1109/TITS.2022.3168232.

11. Kim, E., D. Kim, and K. Sohn. Imputing Qualitative Attributes for Trip Chains Extracted from Smart Card Data Using a Conditional Generative Adversarial Network. *Transportation Research Part C*, Vol. 137, No. February, 2022, p. 103616. https://doi.org/10.1016/j.trc.2022.103616.

12. Theis, L., A. van den Oord, and M. Bethge. A Note on the Evaluation of Generative Models. *4th International Conference on Learning Representations, ICLR 2016 - Conference Track Proceedings*, 2016, pp. 1–10.

13. Sajjadi, M. S. M., O. Bousquet, O. Bachem, M. Lucic, and S. Gelly. Assessing Generative Models via Precision and Recall. *Advances in Neural Information Processing Systems*, Vol. 2018-Decem, No. Nips, 2018, pp. 5228–5237.

14. Kynkäänniemi, T., T. Karras, S. Laine, J. Lehtinen, and T. Aila. Improved Precision and Recall Metric for Assessing Generative Models. *Advances in Neural Information Processing Systems*, Vol. 32, No. NeurIPS, 2019.

15. Gurumurthy, S., R. K. Sarvadevabhatla, and R. V. Babu. DeLiGAN: Generative Adversarial Networks for Diverse and Limited Data. 2017.

16. Mao, Q., H. Y. Lee, H. Y. Tseng, S. Ma, and M. H. Yang. Mode Seeking Generative Adversarial Networks for Diverse Image Synthesis. *Proceedings of the IEEE Computer Society Conference on Computer Vision and Pattern Recognition*, Vol. 2019-June, 2019, pp. 1429–1437. https://doi.org/10.1109/CVPR.2019.00152.

17. Wang, S., B. Mo, and J. Zhao. Deep Neural Networks for Choice Analysis: Architecture Design with Alternative-Specific Utility Functions. *Transportation Research Part C: Emerging Technologies*, Vol. 112, No. February, 2020, pp. 234–251. https://doi.org/10.1016/j.trc.2020.01.012.

18. Sifringer, B., V. Lurkin, and A. Alahi. Enhancing Discrete Choice Models with Representation Learning. *Transportation Research Part B: Methodological*, Vol. 140, 2020, pp. 236–261. https://doi.org/10.1016/j.trb.2020.08.006.

19. Devlin, J., M. W. Chang, K. Lee, and K. Toutanova. BERT: Pre-Training of Deep Bidirectional Transformers for Language Understanding. *NAACL HLT 2019 - 2019 Conference of the North American Chapter of the Association for Computational Linguistics: Human Language Technologies - Proceedings of the Conference*, Vol. 1, No. Mlm, 2019, pp. 4171–4186.

20. Kingma, D. P., and M. Welling. An Introduction to Variational Autoencoders. 2019. https://doi.org/10.1561/2200000056.

21. Goodfellow, I. J., J. Pouget-Abadie, M. Mirza, B. Xu, D. Warde-Farley, S. Ozair, A. Courville, and Y. Bengio. Generative Adversarial Networks. 2014.

22. Gulrajani, I., F. Ahmed, M. Arjovsky, V. Dumoulin, and A. Courville. Improved Training of Wasserstein GANs. 2017.

23. Sun, L., A. Erath, and M. Cai. A Hierarchical Mixture Modeling Framework for Population Synthesis. *Transportation Research Part B: Methodological*, Vol. 114, 2018, pp. 199–212. https://doi.org/10.1016/j.trb.2018.06.002.

24. Naeem, M. F., S. J. Oh, Y. Uh, Y. Choi, and J. Yoo. Reliable Fidelity and Diversity Metrics for Generative Models. *37th International Conference on Machine Learning, ICML 2020*, Vol. PartF16814, 2020, pp. 7133–7142.

25. Tsamardinos, I., L. E. Brown, and C. F. Aliferis. The Max-Min Hill-Climbing Bayesian Network Structure Learning Algorithm. *Machine Learning*, Vol. 65, No. 1, 2006, pp. 31–78. https://doi.org/10.1007/s10994-006-6889-7.

26. Miller, E. J. Modeling the Demand for New Transportation Services and Technologies. *Transportation Research Record*, Vol. 2658, No. 2658, 2017, pp. 1–7. https://doi.org/10.3141/2658-01.

27.   Pereira, R. H. M., T. Schwanen, and D. Banister. Distributive Justice and Equity in Transportation. *Transport Reviews*, Vol. 37, No. 2, 2017, pp. 170–191. https://doi.org/10.1080/01441647.2016.1257660.

28.   Zheng, Y., S. Wang, and J. Zhao. Equality of Opportunity in Travel Behavior Prediction with Deep Neural Networks and Discrete Choice Models. *Transportation Research Part C: Emerging Technologies*, Vol. 132, 2021. https://doi.org/10.1016/j.trc.2021.103410.